\documentclass[a4paper]{article}

\usepackage{INTERSPEECH2019}
\usepackage{xcolor}
\usepackage{hyperref}
\usepackage{upgreek}
\usepackage{siunitx}

\graphicspath{{Figures/}}

\newcommand{\eg}{\textit{e}.\textit{g}. }
\newcommand{\R}{\mathbb{R}}
\newcommand{\T}{^\mathsf{T}}

\renewcommand{\vec}[1]{\mathbf{#1}}
\renewcommand{\mat}[1]{\mathbf{#1}}
\renewcommand{\mu}{\bm{\upmu}}
\renewcommand{\omega}{\bm{\upomega}}
\renewcommand{\psi}{\bm{\uppsi}}
\renewcommand{\phi}{\bm{\upphi}}
\renewcommand{\epsilon}{\bm{\upepsilon}}

\title{Unleashing the Unused Potential of I-Vectors Enabled by GPU Acceleration}

\name{Ville Vestman$^{1,2}$, Kong Aik Lee$^1$, Tomi H. Kinnunen$^2$, Takafumi Koshinaka$^1$}

\address{
  $^1$Biometrics Research Laboratories, NEC Corporation, Japan\\
  $^2$Computational Speech Group, University of Eastern Finland, Finland}
\email{vvestman@cs.uef.fi, k-lee@ax.jp.nec.com, tkinnu@cs.uef.fi, koshinak@ap.jp.nec.com}

\begin{document}

\setlength{\floatsep}{0pt plus 2.0pt minus 2.0pt}
\setlength{\textfloatsep}{8pt plus 2.0pt minus 2.0pt}
\setlength{\dbltextfloatsep}{10pt plus 2.0pt minus 2.0pt}
\setlength{\intextsep}{6pt plus 2.0pt minus 2.0pt}

\setlength{\abovedisplayskip}{4pt plus 1.0pt minus 1.0pt}
\setlength{\belowdisplayskip}{4pt plus 1.0pt minus 1.0pt}
\setlength{\abovedisplayshortskip}{-2pt plus 1.0pt minus 1.0pt}
\setlength{\belowdisplayshortskip}{2pt plus 1.0pt minus 1.0pt}

\maketitle
\begin{abstract}
\vspace{-1mm}
Speaker embeddings are continuous-value vector representations that allow easy comparison between voices of speakers with simple geometric operations. Among others, i-vector and x-vector have emerged as the mainstream methods for speaker embedding. In this paper, we illustrate the use of modern computation platform to harness the benefit of GPU acceleration for i-vector extraction. In particular, we achieve an acceleration of $3000$ times in frame posterior computation compared to real time and $25$ times in training the i-vector extractor compared to the CPU baseline from Kaldi toolkit. This significant speed-up allows the exploration of ideas that were hitherto impossible. In particular, we show that it is beneficial to update the universal background model (UBM) and re-compute frame alignments while training the i-vector extractor. Additionally, we are able to study different variations of i-vector extractors more rigorously than before. In this process, we reveal some undocumented details of Kaldi's i-vector extractor and show that it outperforms the standard formulation by a margin of 1 to 2\% when tested with VoxCeleb speaker verification protocol. All of our findings are asserted by ensemble averaging the results from multiple runs with random start.

\end{abstract}

\vspace{-2mm}
\section{Introduction}
\vspace{-1mm}

A decade ago, the \emph{i-vector} speaker embedding was introduced~\cite{dehak2009discriminative}. Since its introduction, it has remained as a standard solution for speaker recognition until recent years when it was excelled in many tasks by the deep neural network based embeddings \cite{snyder2017deep, snyder2018spoken}. The recent developments are a result of the widespread interest among researchers to adopt deep learning techniques in their research. The most recent rise of deep learning has been partially made possible by the year-by-year increasing computation resources \cite{goodfellow2016deep}, and especially the use of \emph{graphics processing units} (GPUs) to harness the benefits of massive parallelism even with consumer level devices.

While GPUs are heavily adopted in deep learning, they can also be conveniently utilized for the traditional learning of generative models such as the \emph{total variability model} \cite{dehak2011front} underlying i-vector extraction. So far, this has been a largely unexploited possibility despite the fact that full-fledged i-vector extractors tend to be slow to train. The slowness of training has often forced many researchers to limit their experimental validation, for example by limiting the number of training iterations, or by relaying on the results from a single run with random initialization. In addition, simplifications and approximations of the model have been proposed to reduce the computational load~\cite{glembek2011simplification, cumani2014factorized, xu2018generalizing}.

For the current work, we utilize GPU to accelerate i-vector extraction and the total variability model training to alleviate the above limitations. The obtained speed-up allows us to study i-vector extractors in a more detailed manner than what has been possible previously. For example, we can train i-vector extractors \emph{without} any approximations for hundreds of iterations to study the optimal number of iterations to maximize the speaker recognition performance. In addition, we are able to obtain more reliable comparisons between different variations of extractors by averaging the results of multiple runs with different random initializations of the model. For instance, the extractor training can differ in terms of whether model parameters are \emph{re-estimated} using \emph{minimum divergence criterion} \cite{kenny2005joint} and whether the residual covariance matrix of the model is updated.

Further, we re-explore the idea of updating frame alignments during the training of i-vector extractor, which could potentially enhance the model fit and the resulting speaker recognition performance. The idea of updating the alignments was originally presented in the context of \emph{eigenvoice} modeling for automatic speech recognition \cite{kenny2005eigenvoice}, but has received limited attention in the context of i-vectors for speaker recognition. In eigenvoice modeling, the alignment update is performed using speaker-dependent supervectors, which is not suitable approach for speaker recognition as it would tend to model out the speaker information from the i-vectors. Instead, we update the global UBM mean supervector to realign the training data.

In the experiments, we extensively utilize our GPU re-implementation of Kaldi speech recognition toolkit's \cite{povey2011kaldi} \mbox{i-vector} extractor. The implementation in Kaldi has some special traits, which, to the best of our knowledge, have not been extensively documented. Most notably, in Kaldi's implementation, the bias term is augmented to the \emph{total variability matrix}~\cite{dehak2011front}, which causes some changes to the minimum divergence re-estimation step and which also eliminates the need of centralizing Baum-Welch statistics \cite{reynolds2000speaker}. As Kaldi is one of the most popular tools used for the speaker recognition research, we consider it worthwhile to document the main differences of the two formulations in the following sections.

\vspace{-1mm}
\section{I-vector speaker embeddings}
\vspace{-1mm}

We compare two different formulations of the \emph{total variability} approach \cite{dehak2011front} of \emph{joint factor analysis} \cite{kenny2007joint} to extract i-vectors. In the total variability model, all of the variability in utterances is modeled using a single subspace only, without having separate subspaces to model speaker and channel effects.

The first of the formulations is the original formulation~\cite{kenny2005eigenvoice, kenny2012small}, which is commonly adopted in many available speaker recognition toolkits \cite{madikeri2016implementation, larcher2016extensible, sadjadi2013MSR}. The second formulation, implemented in the Kaldi speech recognition toolkit, is inspired by the \emph{subspace Gaussian mixture model} \cite{povey2011subspace}. This formulation differs from the standard one as it augments the bias term of the model to the \emph{factor loading matrix}, which allows estimating the bias term and the factor loading matrix jointly.

Common to both formulations is the use of \emph{Baum-Welch statistics} as defined in \cite{kenny2012small}. In this work, we denote the occupancy statistics, first order statistics, and the second order statistics for the Gaussian component $c$ ($c = 1, 2, \ldots, C$) as $n_c$, $\vec{f}_c$, and $\mat{S}_c$, respectively. To obtain unified presentation for the two formulations, we hereafter assume that the first and second order statistics are centered \cite{kenny2008study} for the standard formulation and \emph{not} centered for the augmented formulation.

\vspace{-1mm}
\subsection{Standard formulation}
\vspace{-1mm}

Following the standard formulation, we model the mean vector of the $c$th Gaussian component of utterance $u$ as
\begin{equation}
    \label{eq:model}
    \mu_c(u) = \vec{m}_c + \mat{T}_c \omega(u),
\end{equation}
where $\vec{m}_c$ is a bias term, matrix $\mat{T}_c$ is a projection matrix, and $\omega(u)$ is a latent vector. The latent vector is shared among all the components and we assume that the prior over latent vectors is standard normal. Further, the covariance matrix of the $c$th Gaussian is modeled as
\begin{equation}
    \label{covariance_update}
    \mat{D}_c(u) = \mat{T}_c \mat{\Phi}(u) \mat{T}_c\T + \mat{\Sigma}_c,
\end{equation}
where $\mat{\Phi}(u)$ is the posterior covariance matrix of the latent vector, and $\mat{\Sigma}_c$ is the residual covariance matrix for component~$c$~\cite{kenny2004speaker}.

The posterior covariance matrix $\mat{\Phi}(u)$ and the mean vector $\phi(u)$ for the latent vector are obtained as 
\begin{align}
    \mat{\Phi}(u) &= \left( \mat{I} + \sum_{c=1}^C  n_c(u) \mat{T}_c\T \mat{\Sigma}_c^{-1} \mat{T}_c \right)^{-1}, \label{posterior_covariance} \\
    \phi(u) &= \mat{\Phi}(u) \left( \vec{p} + \sum_{c=1}^C \mat{T}_c\T \mat{\Sigma}_c^{-1}  \vec{f}_c(u) \right), \label{posterior_mean}
\end{align}
where $\vec{p}$ is the \emph{prior offset}, which is $\vec{0}$ in the standard formulation.

The model is trained iteratively using an EM-algorithm, for which the update formulas for matrices $\mat{T}_c$ and $\mat{\Sigma}_c$ are given in~\cite{kenny2005eigenvoice}. In the beginning of training, the matrices $\mat{T}_c$ are initialized with random values drawn from the standard normal distribution. The initial bias terms $\vec{m}_c$ and the residual covariance matrices $\mat{\Sigma}_c$ are obtained as the means and covariances from \emph{universal background model} (UBM) \cite{reynolds2000speaker}. As the training progresses, the residual covariances get smaller as the first term of right-hand side of \eqref{covariance_update} starts to explain parts of the covariance structures of training utterances. 

\vspace{-1mm}
\subsection{Augmented formulation}
\vspace{-1mm}
In the second formulation, we augment the bias terms $\vec{m}_c$ into the matrices $\mat{T}_c$. This is done by assuming non-zero mean for the prior over the first elements of the latent vectors. Then, equation \eqref{eq:model} becomes
\begin{equation}
    \mu_c(u) = \mat{T}_c \omega(u), 
\end{equation}
where $\omega \sim \mathcal{N}(\vec{p}, \mat{I})$ with $\vec{p} = \begin{bmatrix} p & 0 & \cdots & 0 \end{bmatrix}\T$, $p \in \R$.

Assuming that the Baum-Welch statistics are not centralized, the equations \eqref{posterior_covariance} and \eqref{posterior_mean} hold also for the augmented formulation. The EM update equations presented in \cite{kenny2005eigenvoice} remain the same as well\footnote{Although the residual covariance update implemented in Kaldi might seem different than in \cite{kenny2005eigenvoice}, they can be shown to be equivalent.}. It is worth to note that because of the augmentation, the update of matrices $\mat{T}_c$ also updates the bias terms, which reside in the first columns of matrices $\mat{T}_c$.

The model initialization differs slightly from the standard formulation. First, we set $p = 100$ (same as in the Kaldi implementation) and then we fill the first columns of the randomly initialized matrices $\mat{T}_c$ with the values from the mean vectors of the UBM divided by $p$.

\vspace{-1mm}
\section{Training enhancements}
\vspace{-1mm}
The update step of the model training can have many variations. The most basic one is to only update matrices $\mat{T}_c$, while also updating residual covariances $\mat{\Sigma}_c$ gives a slight improvement to the performance as we will demonstrate later. Another way to improve the model is to apply \emph{minimum divergence re-estimation} to make the empirical distribution of i-vectors close to standard normal \cite{kenny2005joint, kenny2012small}. The minimum divergence re-estimation is not quite as straightforward for the augmented formulation as for the standard one. To the best of our knowledge, the procedure for the augmented formulation is not documented elsewhere than in the source code comments of Kaldi, hence we will provide the key details in the following. Finally, further improvements can be obtained by realigning the training data during the training using the updated models.

\vspace{-1mm}
\subsection{Minimum divergence re-estimation}
\vspace{-1mm}
For the minimum divergence re-estimation, we accumulate the sums
\begin{align}
    \vec{h} &= \frac{1}{U} \sum_{u=1}^U \vec{\phi}(u), \\
    \mat{H} &= \frac{1}{U} \sum_{u=1}^U \left[\mat{\Phi}(u) + \vec{\phi}(u) \vec{\phi}(u)\T\right],
\end{align}
during the E-step. Then, a whitening matrix can be computed via \emph{eigendecomposition} (alternatively, via \emph{Cholesky decomposition}) of the covariance matrix $\mat{G} = \mat{H} - \vec{h} \vec{h}^T$. That is, if $\mat{G} = \mat{Q} \mat{\Lambda} \mat{Q}\T$ is an eigendecomposition of $\mat{G}$, then the whitening transform is obtained as $\mat{P}_1 = \mat{\Lambda}^{-\frac{1}{2}} \mat{Q}\T$. Now, the update $\mat{T}_c^{\textrm{upd}} = \mat{T}_c \mat{P}_1^{-1}$, has an effect of whitening the training i-vectors.

In the standard formulation, the above update is sufficient for the minimum divergence estimation. In the augmented formulation, however, we need to apply another transform $\mat{P}_2$ to the matrices $\mat{T}_c^{\textrm{upd}}$ to conform to the prior offset assumption. In specific, after transforming i-vectors with $\mat{P}_1$ and $\mat{P}_2$, they should remain whitened and only the first element (prior offset) of the projected mean vector $\mat{P}_2\mat{P}_1\vec{h}$ should be non-zero.

One option for a transform that can satisfy the requirements set for $\mat{P}_2$ is a reflection about a hyperplane that goes through the origin. This type of transform is known as the \emph{Householder transform} \cite{householder1958unitary}. The Householder transform with a reflection hyperplane that is orthogonal to an unit length vector $\vec{a}$ is defined as
\begin{equation}
\mat{P}_2 = \mat{I} - 2 \vec{a}\vec{a}\T.
\end{equation}
Now, the problem is to find $\vec{a}$ so that the projected mean vector is a scalar multiple of a unit vector $\vec{e}_1 = \begin{bmatrix} 1 & 0 & \cdots & 0 \end{bmatrix}$. That is,
\begin{equation}
\mat{P}_2\mat{P}_1\vec{h} = b \vec{e}_1, \quad b \in \R.
\end{equation}
It can be shown that one solution is
\begin{equation}
\vec{a} = \alpha \vec{\tilde h} + \beta \vec{e}_1,
\end{equation}
where $\vec{\tilde h}$ is $\mat{P}_1\vec{h}$ normalized to unit length $(\vec{\tilde h} = \mat{P}_1\vec{h} / ||\mat{P}_1\vec{h}||)$ and
\begin{equation}
\begin{cases}
    \alpha = \frac{1}{\sqrt{2 (1 - \vec{\tilde h}[1])}} \\
    \beta = -\alpha,
\end{cases}
\end{equation}
where $\vec{\tilde h}[1]$ is the first element of  $\vec{\tilde h}$.

Now, the update $\mat{T}_c^{\textrm{upd}} = \mat{T}_c \mat{P}_1^{-1} \mat{P}_2^{-1}$ whitens and centers the training i-vectors with respect to the prior offset. Finally, the prior offset $\vec{p}$ is updated as follows:
\begin{equation}
    \label{prior_update}
    \vec{p} = \mat{P}_2\mat{P}_1\vec{h}.
\end{equation}

\vspace{-1mm}
\subsection{Realignment of training data}
\vspace{-1mm}
To compute the Baum-Welch statistics used in training, the frames of training utterances are first aligned to the components of the UBM by computing frame posterior probabilities. The posteriors and the Baum-Welch statistics are typically held constant during the training of i-vector extractor.

In \cite{kenny2005eigenvoice}, the frame alignments of the training utterances are updated during the training of factor analysis model for \emph{automatic speech recognition} (ASR). The realignment is done per speaker basis using adapted GMM means and covariances. In the application of speaker recognition, however, this would be counterproductive as it would reduce the amount of speaker information in the latent vectors. What we propose instead, is updating the UBM means with the updated bias terms $\vec{m}_c$ and then using the updated UBM to realign the data, which can potentially lead to a better model fit. To obtain the updated bias terms from the augmented formulation, we simply take the first columns of matrices $\mat{T}_c$ and multiply them with $p$.

In summary, the augmented model with posterior updates is trained by iterating over the following five steps:
\begin{enumerate}
    \item The computation of frame alignments and Baum-Welch statistics using the current UBM \cite{reynolds2000speaker, kenny2012small}.
    \item \textbf{E-step:} The computation of posterior means and covariances for the latent vectors using \eqref{posterior_covariance} and \eqref{posterior_mean} to accumulate the required terms for the M-step.
    \item \textbf{M-step:} The update of matrices $\mat{T}_c$ followed by the update of residual covariances $\mat{\Sigma}_c$ \cite{kenny2005eigenvoice}. 
    \item \textbf{Minimum divergence re-estimation:} The update of matrices $\mat{T}_c$ using the transforms $\mat{P}_1$ and $\mat{P}_2$ followed by the update of the prior offset $\vec{p}$ using \eqref{prior_update}.
    \item If not the last iteration, the update of the mean vectors of the UBM with the first columns of matrices $\mat{T}_c^{\textrm{upd}}$ multiplied by $p$.
\end{enumerate}
After the model has been trained, the updated UBM is used in the testing phase to compute the frame posteriors.

\vspace{-1mm}
\section{Experiments}

\subsection{Experimentation setup}
\vspace{-1mm}
\label{sec:setup}

We built the acoustic front-end of our systems on the basis of Kaldi \cite{povey2011kaldi} i-vector recipe for VoxCeleb \cite{nagrani2017voxceleb, Chung18b}. That is, we relied on Kaldi to extract MFCCs, to perform voice activity detection (VAD), and to train the UBM. We used the same settings as in the Kaldi recipe: The MFCC vectors are $72$-dimensional including delta and double-delta coefficients, and the UBM consists of $2048$ components with full covariance matrices.

Following the Kaldi recipe, the UBM was trained using all of the data from the training parts of VoxCeleb1 and Voxceleb2 consisting of $\num[group-separator={\text{\,}}]{1277344}$ utterances from $7325$ speakers. The i-vector extractors were trained using the $\num[group-separator={\text{\,}}]{100000}$ longest utterances. To train the scoring back-end, the Kaldi recipe uses the whole training data, while we utilized only the VoxCeleb1 proportion to speed up the experimentation. Although this reduced the number of training speakers from $7325$ to $1211$, we did not observe degradation in speaker verification performance\footnote{This might be explained by the fact that VoxCeleb1 has more reliable speaker labels than VoxCeleb2 \cite{Chung18b}.}.

After the i-vector extraction, we centered and length normalized the i-vectors. In addition, if minimum divergence re-estimation was not used, we also whitened the i-vectors before length normalization. Then, we reduced the dimensionality of i-vectors from $400$ to $200$ using \emph{linear discriminant analysis} (LDA) before subjecting them to \emph{probabilistic linear discriminant analysis} (PLDA) scoring \cite{garcia2011analysis}. For testing, we used adopted the VoxCeleb1 speaker verification protocol, which consists of $\num[group-separator={\text{\,}}]{37720}$ trials with an equal number of target and non-target trials.

We ran the experiments on a server having Intel Xeon Gold 6152 CPU with $22$ physical cores and NVIDIA Titan V GPU with $12$ GB of memory. The file I/O operations were performed on a solid-state drive (SSD).

\vspace{-1mm}
\subsection{GPU implementation}
\vspace{-1mm}

\begin{figure}[t]
  \centering
  \includegraphics[trim={4.8cm 5.3cm 10.1cm 2.1cm},clip, width=0.9\linewidth]{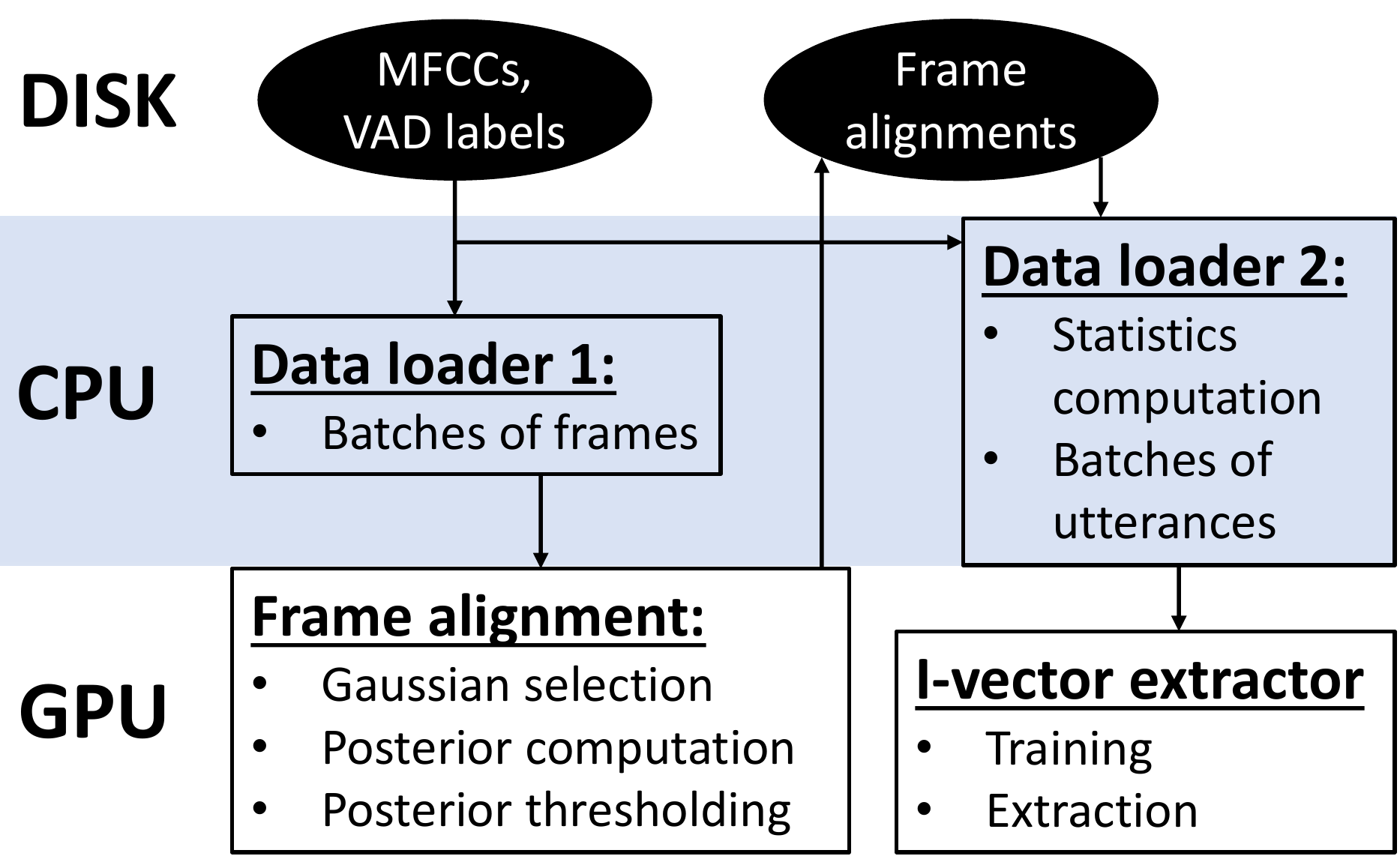}
  \caption{An overview of computational flow of frame alignment, i-vector extraction, and model training using a GPU. 
  To keep the GPU memory requirements constant, fixed size batches of frames and utterances are used for frame alignment and \mbox{i-vector} extraction, respectively.}
  \label{fig:diagram}
\end{figure}

In our implementations of frame alignment and i-vector extraction, we utilized PyTorch \cite{paszke2017automatic} for GPU computations, SciPy ecosystem \cite{scipy} for computations in CPU, and PyKaldi \cite{pykaldi} for reading files stored in Kaldi format. The implementations use multiple CPU cores in parallel as data loaders, which load, preprocess, and feed the data to the GPU (Figure~\ref{fig:diagram}). The data loaders function in parallel with respect to the GPU to keep the GPU utilized all the time.

For frame alignment, we use the same strategy as in Kaldi: First, to reduce the computational load, we use a UBM with diagonal covariance matrices to select the top-20 Gaussian components with the highest frame posteriors for each frame. Second, we compute the posteriors with only the selected components using a full covariance UBM. Finally, we discard the posteriors that are less than $0.025$ and we linearly scale the remaining posteriors so that their sum equals to one. As a result, on average, only four Gaussian indices and the corresponding posteriors are stored to disk per frame.

The Baum-Welch statistics used in i-vector extractor training are computed in CPU, while the rest of the computation is done in GPU. The reason to compute statistics in CPU is as follows: For i-vector extraction implementation, it is natural to feed data in batches of utterances, and statistics provide a fixed size representation of utterances unlike the acoustic features. We opted not to compute statistics beforehand as the disk usage would be excessive; instead we recompute them during each iteration of i-vector extractor training.

With the settings laid out in Section \ref{sec:setup}, the GPU memory usage for alignment computation is about $2.5$ GB and for i-vector extractor training about $4$ GB. The frame alignment can be done about $\num[group-separator={\text{\,}}]{3000}$ times faster than real time (including I/O operations), and assuming that the alignments are ready in the disk, i-vectors can be extracted $\num[group-separator={\text{\,}}]{10000}$ times faster than real time. By using the GPU re-implementation of Kaldi's i-vector extractor training, we were able to obtain $25$-fold reduction in the training times. This number was obtained by training both our GPU implementation and Kaldi's CPU implementation for five iterations and measuring the elapsed times. The training using Kaldi utilized all the available CPU cores in the server.

\begin{figure}[t]
  \centering
  \includegraphics[width=\linewidth]{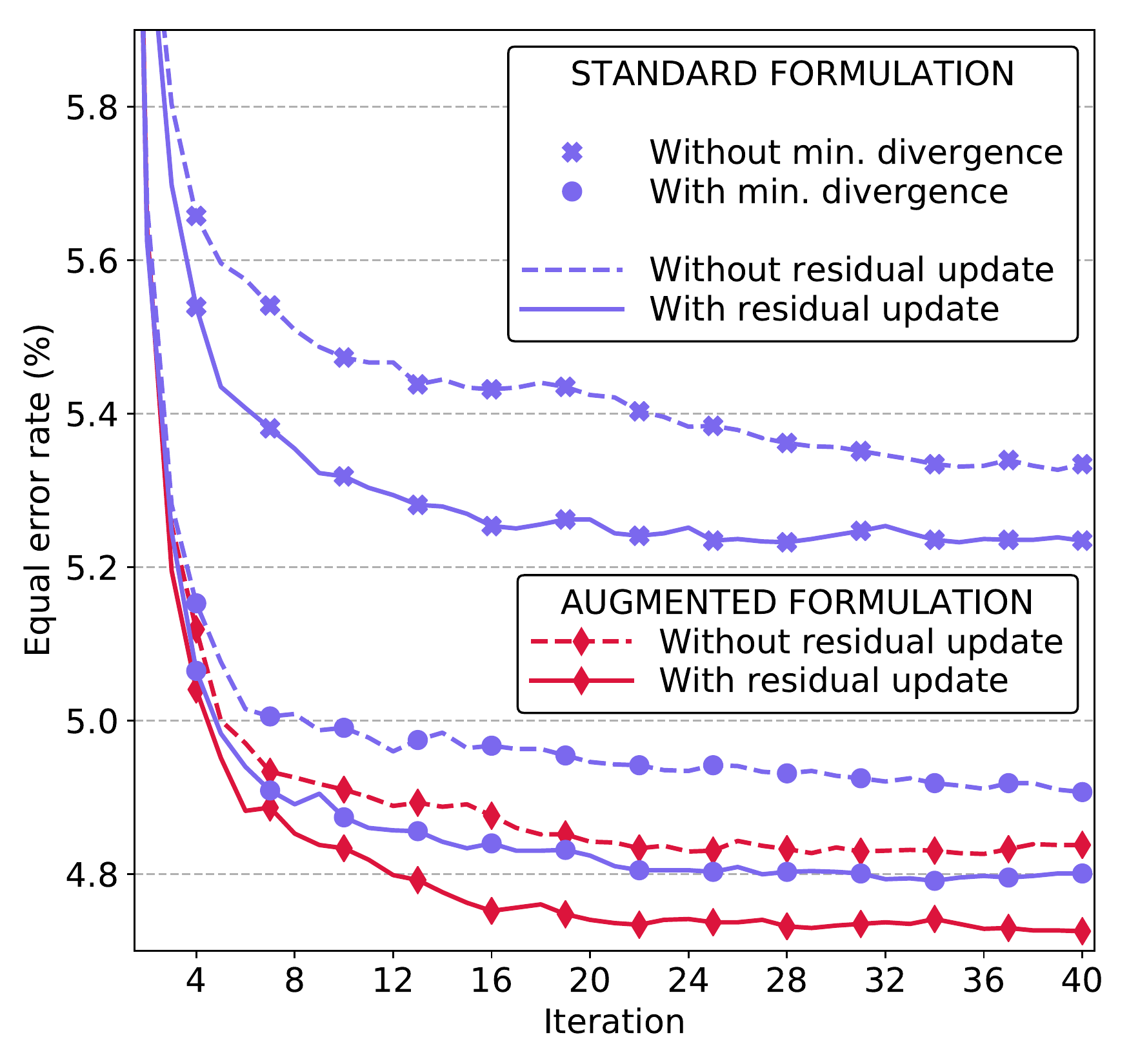}
  \vspace{-6mm}
  \caption{System performance as function of number of iterations in i-vector extractor training. The frame alignments are not updated during the training. The standard formulation has four training variations, which are obtained by either performing or not performing minimum divergence re-estimation and either updating or not updating residual covariance matrices. In the augmented formulation, the minimum divergence re-estimation is always applied. Each curve is obtained as an average of five runs with different random initial values of $\mat{T}_c$.}
  \label{fig:variations}
\end{figure}

\vspace{-1mm}
\subsection{Speaker verification results}
\vspace{-1mm}
We began the experiments by comparing different variations of i-vector extractors to select the best one for further experiments with frame alignment updates. The results of the comparison are shown in Figure \ref{fig:variations}. We observe the following: First, the minimum divergence re-estimation to update the model hyperparameters results in $7.5$ -- $9$\% relative reduction in terms of equal error rate (EER). Second, the update of residual covariance matrices leads to $1.5$ -- $3$\% relative reduction of error rates. Third, the augmented formulations obtain $1$ -- $2$\% lower error rates (relative) than the standard formulations.
Finally, we assert that $22$ iterations are enough to reach the optimal speaker verification performance with the best performing extractors. As our results are averages of five runs, individual runs may converge faster than that. In addition, we confirmed that our assertion is correct by training the augmented model once for $200$ iterations.

\begin{figure}[t]
  \centering
  \includegraphics[width=\linewidth]{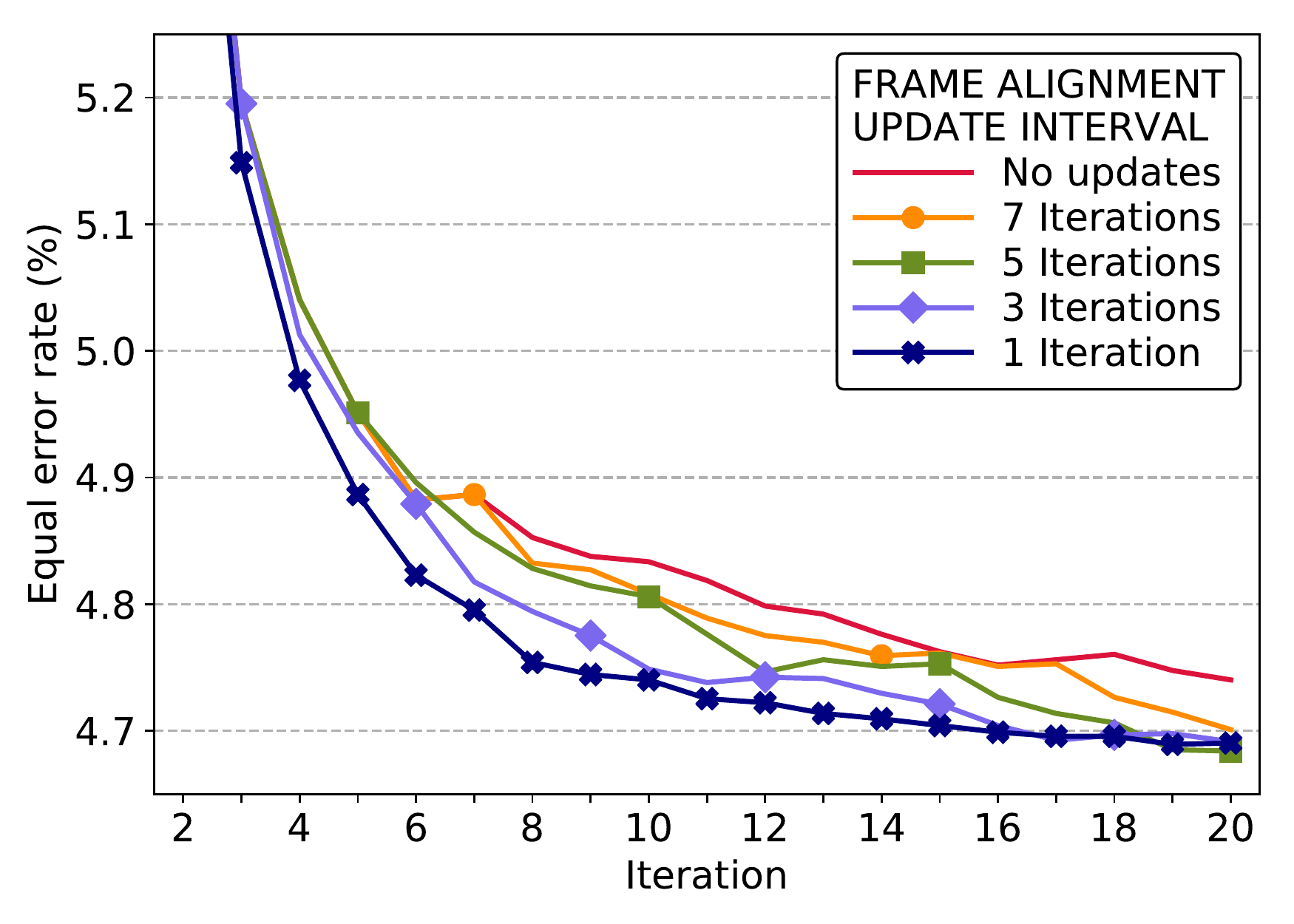}
  \vspace{-6mm}
  \caption{Performance of the augmented formulation for varying intervals of frame alignment updates. The more often the alignments are updated, the faster the system performance improves. Each curve is obtained as an average of five runs with different random initializations.}
  \label{fig:updates}
\end{figure}

Based on the first experiment, we continued to experiment with the realignment of training data using the augmented formulation with residual covariance matrix updates. We varied the interval between the frame posterior updates ranging from updating on every iteration to updating only on every seventh iteration. We display the results in Figure \ref{fig:updates}. The findings are two-fold: First, the more frequently the frame posteriors are updated, the faster the performance improves. Second, updating the posteriors, no matter how frequently, leads to about $1$\% lower error rates (relative) compared to training without updates.

At best, we obtained an EER of $4.6$\%, which could be possibly made closer to $4.0$\% by carefully optimizing configurations in various parts of the system. For comparison, the state-of-the-art system, using x-vectors, obtains EER of $3.1$\% (reported in the Kaldi recipe). This is an expected performance difference between the i-vector and x-vector systems \cite{snyder2018x}.

\vspace{-1mm}
\section{Discussion and conclusions}
\vspace{-1mm}
We have a couple of remarks from the practical aspect of the study. First, we found that by using the modern deep learning platforms, such as PyTorch, the implementation of GPU accelerated algorithms for generative models is almost as straightforward as it is with their non-GPU counterparts (\eg NumPy). The only concern is the limited amount of memory in GPUs. This limitation can be often circumvented by relying on the computational power of GPUs to recompute values that do not fit into the memory.

The second remark concerns the update of the UBM means using the bias terms $\vec{m}_c$ of the model. For this purpose, we only used the augmented formulation, but it can be done also with the standard formulation by updating the means in the minimum divergence step using a formula $\vec{m}_c^\textrm{upd} = \vec{m}_c + \mat{T}_c \vec{h}$ \cite{kenny2008study}. However, we found that updating the means in this way did not work well together with residual covariance updates.

In summary, the results of the study showed that the choice of the training algorithm for i-vector extractor matters as the relative change in equal error rate between the worst and the best variations was $11.4$\%. For the optimal performance, our recommendation is to use the augmented formulation including the residual covariance updates and the updates of frame alignments. Additionally we found that the extractors reach their maximum performance after $22$ training iterations.

\vspace{-1mm}
\section{Acknowledgements}
\vspace{-1mm}
This work was partially supported by Academy of Finland (proj. \#309629) and by the Doctoral Programme in Science, Technology and Computing (SCITECO) of the UEF. The authors at UEF were also supported by NVIDIA Corporation with the donation of Titan V GPU.

\bibliographystyle{IEEEtran}
\bibliography{mybib}

\begin{thebibliography}{10}
\providecommand{\url}[1]{#1}
\csname url@samestyle\endcsname
\providecommand{\newblock}{\relax}
\providecommand{\bibinfo}[2]{#2}
\providecommand{\BIBentrySTDinterwordspacing}{\spaceskip=0pt\relax}
\providecommand{\BIBentryALTinterwordstretchfactor}{4}
\providecommand{\BIBentryALTinterwordspacing}{\spaceskip=\fontdimen2\font plus
\BIBentryALTinterwordstretchfactor\fontdimen3\font minus
  \fontdimen4\font\relax}
\providecommand{\BIBforeignlanguage}[2]{{%
\expandafter\ifx\csname l@#1\endcsname\relax
\typeout{** WARNING: IEEEtran.bst: No hyphenation pattern has been}%
\typeout{** loaded for the language `#1'. Using the pattern for}%
\typeout{** the default language instead.}%
\else
\language=\csname l@#1\endcsname
\fi
#2}}
\providecommand{\BIBdecl}{\relax}
\BIBdecl

\bibitem{dehak2009discriminative}
N.~Dehak, ``Discriminative and generative approaches for long-and short-term
  speaker characteristics modeling: application to speaker verification,''
  Ph.D. dissertation, {\'E}cole de technologie sup{\'e}rieure, 2009.

\bibitem{snyder2017deep}
D.~Snyder, D.~Garcia-Romero, D.~Povey, and S.~Khudanpur, ``Deep neural network
  embeddings for text-independent speaker verification.'' in
  \emph{Interspeech}, 2017, pp. 999--1003.

\bibitem{snyder2018spoken}
\BIBentryALTinterwordspacing
D.~Snyder, D.~Garcia-Romero, A.~McCree, G.~Sell, D.~Povey, and S.~Khudanpur,
  ``Spoken language recognition using x-vectors,'' in \emph{Proc. Odyssey 2018
  The Speaker and Language Recognition Workshop}, 2018, pp. 105--111. [Online].
  Available: \url{http://dx.doi.org/10.21437/Odyssey.2018-15}
\BIBentrySTDinterwordspacing

\bibitem{goodfellow2016deep}
I.~Goodfellow, Y.~Bengio, and A.~Courville, \emph{Deep learning}.\hskip 1em
  plus 0.5em minus 0.4em\relax MIT press, 2016.

\bibitem{dehak2011front}
N.~Dehak, P.~J. Kenny, R.~Dehak, P.~Dumouchel, and P.~Ouellet, ``Front-end
  factor analysis for speaker verification,'' \emph{IEEE Transactions on Audio,
  Speech, and Language Processing}, vol.~19, no.~4, pp. 788--798, 2011.

\bibitem{glembek2011simplification}
O.~Glembek, L.~Burget, P.~Mat{\v{e}}jka, M.~Karafi{\'a}t, and P.~Kenny,
  ``Simplification and optimization of i-vector extraction,'' in \emph{2011
  IEEE International Conference on Acoustics, Speech and Signal Processing
  (ICASSP)}.\hskip 1em plus 0.5em minus 0.4em\relax IEEE, 2011, pp. 4516--4519.

\bibitem{cumani2014factorized}
S.~Cumani and P.~Laface, ``Factorized sub-space estimation for fast and memory
  effective i-vector extraction,'' \emph{IEEE/ACM Transactions on Audio,
  Speech, and Language Processing}, vol.~22, no.~1, pp. 248--259, 2014.

\bibitem{xu2018generalizing}
L.~Xu, K.~A. Lee, H.~Li, and Z.~Yang, ``Generalizing i-vector estimation for
  rapid speaker recognition,'' \emph{IEEE/ACM Transactions on Audio, Speech and
  Language Processing (TASLP)}, vol.~26, no.~4, pp. 749--759, 2018.

\bibitem{kenny2005joint}
P.~Kenny, ``Joint factor analysis of speaker and session variability: Theory
  and algorithms,'' \emph{CRIM, Montreal,(Report) CRIM-06/08-13}, vol.~14, pp.
  28--29, 2005.

\bibitem{kenny2005eigenvoice}
P.~Kenny, G.~Boulianne, and P.~Dumouchel, ``Eigenvoice modeling with sparse
  training data,'' \emph{IEEE Transactions on Speech and Audio Processing},
  vol.~13, no.~3, pp. 345--354, 2005.

\bibitem{povey2011kaldi}
D.~Povey, A.~Ghoshal, G.~Boulianne, L.~Burget, O.~Glembek, N.~Goel,
  M.~Hannemann, P.~Motlicek, Y.~Qian, P.~Schwarz \emph{et~al.}, ``The {Kaldi}
  speech recognition toolkit,'' IEEE Signal Processing Society, Tech. Rep.,
  2011.

\bibitem{reynolds2000speaker}
D.~A. Reynolds, T.~F. Quatieri, and R.~B. Dunn, ``Speaker verification using
  adapted {Gaussian} mixture models,'' \emph{Digital signal processing},
  vol.~10, no. 1-3, pp. 19--41, 2000.

\bibitem{kenny2007joint}
P.~Kenny, G.~Boulianne, P.~Ouellet, and P.~Dumouchel, ``Joint factor analysis
  versus eigenchannels in speaker recognition,'' \emph{IEEE Transactions on
  Audio, Speech, and Language Processing}, vol.~15, no.~4, pp. 1435--1447,
  2007.

\bibitem{kenny2012small}
P.~Kenny, ``A small footprint i-vector extractor,'' in \emph{Odyssey}, vol.
  2012, 2012, pp. 1--6.

\bibitem{madikeri2016implementation}
S.~Madikeri, S.~Dey, P.~Motlicek, and M.~Ferras, ``Implementation of the
  standard i-vector system for the kaldi speech recognition toolkit,'' Idiap,
  Tech. Rep., 2016.

\bibitem{larcher2016extensible}
A.~Larcher, K.~A. Lee, and S.~Meignier, ``An extensible speaker identification
  {SIDEKIT} in {Python},'' in \emph{2016 IEEE International Conference on
  Acoustics, Speech and Signal Processing (ICASSP)}.\hskip 1em plus 0.5em minus
  0.4em\relax IEEE, 2016, pp. 5095--5099.

\bibitem{sadjadi2013MSR}
S.~O. Sadjadi, M.~Slaney, and L.~P. Heck, ``{MSR} identity toolbox v1.0: A
  {MATLAB} toolbox for speaker recognition research,'' 2013.

\bibitem{povey2011subspace}
D.~Povey, L.~Burget, M.~Agarwal, P.~Akyazi, F.~Kai, A.~Ghoshal, O.~Glembek,
  N.~Goel, M.~Karafi{\'a}t, A.~Rastrow \emph{et~al.}, ``The subspace {Gaussian}
  mixture model---{A} structured model for speech recognition,'' \emph{Computer
  Speech \& Language}, vol.~25, no.~2, pp. 404--439, 2011.

\bibitem{kenny2008study}
P.~Kenny, P.~Ouellet, N.~Dehak, V.~Gupta, and P.~Dumouchel, ``A study of
  interspeaker variability in speaker verification,'' \emph{IEEE Transactions
  on Audio, Speech, and Language Processing}, vol.~16, no.~5, pp. 980--988,
  2008.

\bibitem{kenny2004speaker}
P.~Kenny, G.~Boulianne, P.~Ouellet, and P.~Dumouchel, ``Speaker adaptation
  using an eigenphone basis,'' \emph{IEEE transactions on speech and audio
  processing}, vol.~12, no.~6, pp. 579--589, 2004.

\bibitem{householder1958unitary}
A.~S. Householder, ``Unitary triangularization of a nonsymmetric matrix,''
  \emph{Journal of the ACM (JACM)}, vol.~5, no.~4, pp. 339--342, 1958.

\bibitem{nagrani2017voxceleb}
A.~Nagrani, J.~S. Chung, and A.~Zisserman, ``{VoxCeleb}: A large-scale speaker
  identification dataset,'' \emph{Proc. Interspeech 2017}, pp. 2616--2620,
  2017.

\bibitem{Chung18b}
J.~S. Chung, A.~Nagrani, and A.~Zisserman, ``{VoxCeleb2}: Deep speaker
  recognition,'' in \emph{INTERSPEECH}, 2018.

\bibitem{garcia2011analysis}
D.~Garcia-Romero and C.~Y. Espy-Wilson, ``Analysis of i-vector length
  normalization in speaker recognition systems,'' in \emph{Twelfth Annual
  Conference of the International Speech Communication Association}, 2011.

\bibitem{paszke2017automatic}
A.~Paszke, S.~Gross, S.~Chintala, G.~Chanan, E.~Yang, Z.~DeVito, Z.~Lin,
  A.~Desmaison, L.~Antiga, and A.~Lerer, ``Automatic differentiation in
  {PyTorch},'' in \emph{NIPS-W}, 2017.

\bibitem{scipy}
\BIBentryALTinterwordspacing
E.~Jones, T.~Oliphant, P.~Peterson \emph{et~al.}, ``{SciPy}: Open source
  scientific tools for {Python},'' 2001--. [Online]. Available:
  \url{http://www.scipy.org/}
\BIBentrySTDinterwordspacing

\bibitem{pykaldi}
D.~Can, V.~R. Martinez, P.~Papadopoulos, and S.~S. Narayanan, ``Pykaldi: A
  {Python} wrapper for {Kaldi},'' in \emph{IEEE International Conference on
  Acoustics, Speech and Signal Processing (ICASSP)}.\hskip 1em plus 0.5em minus
  0.4em\relax IEEE, 2018.

\bibitem{snyder2018x}
D.~Snyder, D.~Garcia-Romero, G.~Sell, D.~Povey, and S.~Khudanpur, ``X-vectors:
  Robust {DNN} embeddings for speaker recognition,'' in \emph{2018 IEEE
  International Conference on Acoustics, Speech and Signal Processing
  (ICASSP)}.\hskip 1em plus 0.5em minus 0.4em\relax IEEE, 2018, pp. 5329--5333.

\end{thebibliography}

\end{document}